\documentclass[letterpaper, 10 pt, conference]{ieeeconf}  

\IEEEoverridecommandlockouts                              

\usepackage{stfloats}
\usepackage{float} 

\usepackage{booktabs} 
\usepackage{tabularx} 
\title{\LARGE \bf
Co-STAR: Cognitive Stimulation Therapy by an Autonomous Robot for Dementia - A One-Week In-Home Study
}

\author{Emmanuel Akinrintoyo and Nicole Salomons 
\thanks{Emmanuel Akinrintoyo and Nicole Salomons are with  Imperial College London
        {\tt\small ,  e.akinrintoyo23@imperial.ac.uk,
n.salomons@imperial.ac.uk}}%
        }%

\usepackage{graphicx}
\usepackage{dirtytalk}
\usepackage{booktabs} 
\usepackage{amsmath}

\begin{document}

\maketitle
\thispagestyle{empty}
\pagestyle{empty}

\begin{abstract}
Cognitive therapies have been shown to enhance the quality of life and well-being of people living with dementia (PwDs). However, their use remains limited due to a shortage of trained professionals and the significant time and training required of informal caregivers. To address this gap, we developed and deployed a social robot capable of autonomously delivering cognitive stimulation therapy (CST) in the home. Nine PwDs participated in a one-week ($7$ days) study that involved daily robot-led sessions. Participants engaged positively with the system, completing nearly half of the scheduled sessions, an adherence rate higher than typically observed in caregiver-led CST. Our findings also highlight the crucial role of family members, who often supported session initiation and occasionally joined the activities, enriching the interactions. This work demonstrates the feasibility and potential of socially assistive robots to deliver in-home cognitive therapy, offering a scalable approach to extend access to dementia care.
\end{abstract}

\section{INTRODUCTION}

Cognitive Stimulation Therapy (CST)~\cite{woods2006improved, spector2010cognitive} is an evidence-based intervention considered the most effective and widely endorsed approach for dementia. CST is the leading non-drug intervention recommended by dementia care guidelines in countries such as the UK and Canada~\cite{duff2018dementia, vedel2020cccdtd5}. Randomised control trials of CST show improvements in cognition and quality of life~\cite{woods2006improved, woods2023cognitive}. Meta-analyses conclude that it yields reliable and meaningful cognitive gains. This gain is roughly equivalent to a six-month delay in cognitive decline typically expected in mild–to-moderate dementia~\cite{woods2023cognitive}.

Individualised CST (iCST) is a personalised variant of CST~\cite{yates2015development,orrell2017impact}. It is delivered one-to-one to a PwD by either a professional or a family member. 


Despite the potential benefits of iCST, a large multicentre trial reported \emph{low adherence}. Only \(\mathbf{40\%}\) of iCST dyads achieved \(\geq 2\) sessions per week despite a prescription of three sessions weekly~\cite{orrell2017impact}. Qualitative and field studies indicate that many dyads struggle to maintain consistency due to caregivers’ limited time, competing responsibilities, stress, and lack of training~\cite{leung2017experiences,yates2016field,orrell2017impact}. As a result, suboptimal adherence limits the practical effectiveness of iCST~\cite{yates2016field}, revealing a persistent gap between the proven efficacy of CST in controlled settings and the realities of in-home delivery.

\begin{figure}[t] 
    \centering
    \includegraphics[width=0.38\textwidth]{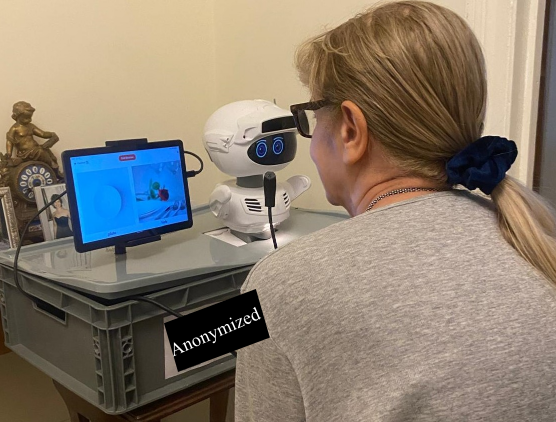}
    \caption{A participant completing a cognitive stimulation therapy session in their living room with Co-STAR.}
    \label{fig:img1}
\end{figure}

Social Assistive Robots (SARs) are robots designed to support and assist individuals through social interaction~\cite{scassellati2018improving, yu2022socially}. They can effectively deliver structured interactions consistently. This makes them suitable for dementia interventions where engagement is essential~\cite{abdi2018scoping}. SARs can also personalise to the PwDs' interests, needs and preferences~\cite{abdollahi2022artificial}. Unlike human caregivers, they do not show frustration and impatience during repeated interactions~\cite{leung2017experiences, abdollahi2022artificial}. 


In this paper, we present a novel robotic system that provides personalised iCST therapy to PwDs in their homes. The robotics system delivered multimodal interaction through speech and a tablet-based visual interface (see Figure \ref{fig:img1}). It delivered activities geared towards supporting memory recall, reasoning, and reminiscence. We recruited nine PwDs who interacted with our system daily for a week in their homes. Participants initiated the sessions on demand via the tablet, allowing the therapy to fit around daily routines. 

This paper makes two novel contributions. First, to the best of our knowledge, this is the first in-home deployment of an autonomous social robot delivering the evidence-based CST for dementia care. To date, most in-home autonomous robots for dementia offer generic stimulation (games, reminders, chat) rather than structured, evidence-based therapy. Second, we provide an in-depth examination of factors that shape the success of these systems for PwDs. We capture nuanced behavioural and contextual insights into what drives multi-session interaction engagement, the role of family members and friends in supporting use, and factors that influence adherence. Beyond dementia care, this work informs SAR design for home-based rehabilitation, mental health, and chronic care. 


\section{BACKGROUND}
This section overviews iCST as an effective intervention.

\subsection{Individual Cognitive Stimulation Therapy}

iCST builds on evidence from traditional cognitive stimulation, which has been found to significantly improve well-being, problem-solving and memory~\cite{yates2015development}. Notably, iCST broadens the reach and inclusivity of cognitive and social stimulation for older persons unable or unwilling to attend group sessions~\cite{yates2015development}. Besides the benefits to PwDs, iCST has been found to enhance caregiver well-being, while strengthening the caregiver–patient relationship~\cite{orrell2017impact}.

iCST generally involves three 45-minute weekly sessions. The sessions are structured, conversational, and person-centred. iCST activities provide mental stimulation focusing on everyday themes such as popular places, sayings, hobbies, famous faces, and daily living activities~\cite{orrell2017impact}. These activities evoke positive emotions. Instead of factual quizzing, they help PwDs reminisce about past positive memories. This includes recalling previous travels, familiar faces, and common proverbs. These therapeutic mechanisms emphasise semantic activation, language production, and executive engagement~\cite{yates2015development, orrell2017impact}. They are conducted through guided discussion, light problem-solving, and orientation prompts.

\subsection{Robots for Cognitive Stimulation}
Recent work has explored cognitive support robots. For example, CARMEN~\cite{bouzida2024carmen} deployed an autonomous robot to provide compensatory cognitive training (CCT) for three persons living with mild cognitive impairment (PwMCI). The one-week study demonstrated the feasibility of home deployment. However, it did not evaluate autonomous delivery of iCST. Moreover, it focused on CCT for PwMCI rather than CST for dementia. 

Cruz-Sandoval et al.~\cite{cruz2020social} facilitated a nine-week nursing home study with a group of eight PwDs interacting with their autonomous social robot. It focused on combining music therapy, reminiscence, and cognitive games to address neuropsychiatric symptoms. It found a significant reduction in dementia-related behavioural symptoms. Gasteiger et al.~\cite{gasteiger2021robot} explored robot-delivered cognitive stimulation games with ten community-dwelling older adults in two retirement villages. It was found to be potentially beneficial for improving cognitive functioning. Lee et al.~\cite{lee2020four} conducted a randomised controlled trial (RCT) with $46$ adults aged $60+$ with mild cognitive impairment (MCI), where participants used a home-care robot for $60$-minute cognitive training sessions daily for four weeks. The intervention significantly improved working memory and reduced anxiety. Park et al.~\cite{park2021humanoid} conducted an RCT with $135$ older adults with memory complaints or MCI who completed $60$-minute robot-assisted sessions with the humanoid robot twice weekly for six weeks. It improved overall cognition and reduced depression. Abdollahi et al.~\cite{abdollahi2022artificial} evaluated an emotionally intelligent social robot, Ryan, with $10$ older adults in a senior living facility. It improved mood and reduced depression. Catricalà et al.~\cite{catricala2025exploiting} personalised serious games delivered by a Pepper robot. Fifteen older adults with MCI played memory-based games using their personal memories over $12$ weeks. It found that using personal memories makes cognitive training more engaging.

These prior studies show that robot-delivered games can engage older persons~\cite{gasteiger2021robot, park2021humanoid}, support social interaction~\cite{abdollahi2022artificial,cruz2020social}, personalise content \cite{catricala2025exploiting}, and engage PwDs \cite{cruz2020social,abdollahi2022artificial}. However, only two of these studies~\cite{cruz2020social,abdollahi2022artificial} were conducted with PwDs, whereas most were only tested with older adults without diagnosis or with MCI \cite{lee2020four, gasteiger2021robot, park2021humanoid, catricala2025exploiting, bouzida2024carmen}. Furthermore, studies such as Cruz-Sandoval et al.~\cite{cruz2020social} investigated multi-activity CST programs for groups of PwDs, rather than the one-to-one CST delivery examined in this study. Others focused on games~\cite{gasteiger2021robot,park2021humanoid, catricala2025exploiting} or companionship~\cite{abdollahi2022artificial} rather than a structured therapy protocol. Lastly, most of these studies were done in day centres or residential facilities~\cite{gasteiger2021robot,cruz2020social, park2021humanoid, abdollahi2022artificial}, with only~\cite{bouzida2024carmen} providing in-home autonomous interaction (in MCI). This highlights the novelty of our work, being the only study that was conducted autonomously in the home, for multi-session interactions with PwDs.



\section{System}

In previous work, we consulted with $16$ dementia stakeholders, including formal and informal caregivers~\cite{akinrintoyo2025home}. The consultations revealed the potential of robotics systems to deliver in-home iCST. Design guidelines and recommendations were gathered from the consultations~\cite{akinrintoyo2025home}. They included the structure of the robot-based therapy and the necessary design goals needed for an effective intervention. Subsequently, we designed a robotic system to deliver iCST at home. It included natural speech interaction, multimodal visual prompts, adaptive activity scheduling, and personalised narratives. The system was tested in $30$-minute sessions with care professionals and PwDs. The feedback was used to develop the final version of the system, presented here.

\subsection{iCST Activities}

We designed iCST session activities to stimulate different cognitive processes. They were anchored on seven design principles, obtained from stakeholder consultations in our previous work~\cite{akinrintoyo2025home}: the system encouraged connection rather than correction, focused on positive encouragement rather than detecting errors of the PwD; it provided cognitive stimulation rather than repetitive, drill-and-practice tasks; and privileged opinions over facts to invite reminiscence; autonomy was supported to ensure little human operation was needed; privacy was handled separately via privacy-by-design (local processing); personalisation was central to ensure relevant and engaging interactions were provided; lastly, participants were encouraged to invite their family members to observe and join during the therapy.

\begin{figure*}[t] 
    \centering
    \includegraphics[width=1.0\textwidth]{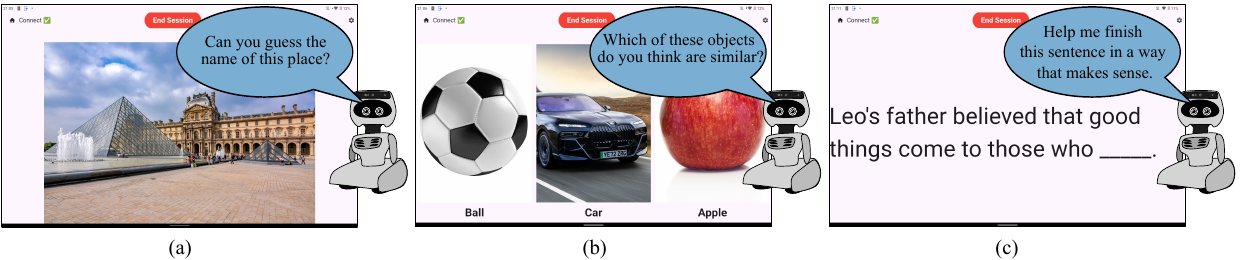} 
    \caption{(a) Popular Places Activity, (b) Object Categorisation Activity, (c) Common Sayings Activity}
    \label{fig:wide2}
\end{figure*}

Here we describe the five developed iCST activities. The Popular Places activity displayed images of well-known landmarks to aid memory recall. Famous Faces involved recognising notable people to strengthen memory associations. Common Sayings involved completing familiar proverbs to promote language activation. In Word Association, a PwD identifies the thematic connection among multiple images to reinforce semantic memory. Lastly, Object Categorisation requires a user to group related objects. Together, the five iCST activities foster memory recall, language stimulation and semantic processing. The Word Association and Common Sayings activities were repeated with different content to complete the seven days of deployment.


Before the deployment day, participants shared some personal details (e.g., jobs, travels, hobbies). The information was then embedded into the interactions for personalisation. For example, we incorporated some of the popular places they had visited into the Popular Places activity. 

\subsection{iCST Session}

Each session was designed to last $30$ minutes. Participants initiated the iCST sessions by pressing the start button on the tablet after unlocking it. The robot then greeted the participant and provided a brief introduction to the day's activity with a fun fact. For each round of the activity, the robot showed pictures on the tablet and asked several questions related to the pictures. In the Sayings activity, it displayed text instead. A text-to-speech system captured the PwD's utterances. We used text matching to identify each answer as correct or incorrect. When the answer was correct, the robot provided brief positive reinforcement (e.g., \say{Great job}); when the answer was incorrect, the robot offered supportive feedback (e.g., \say{You’re close}, \say{Almost there}) and then provided the correct answer. At the end of the activity, the robot gave a brief wrap-up, thanked the participant, and highlighted one or two things they discussed.

\subsection{Hardware}
\begin{figure}[t] 
    \centering
    \includegraphics[width=0.41\textwidth]{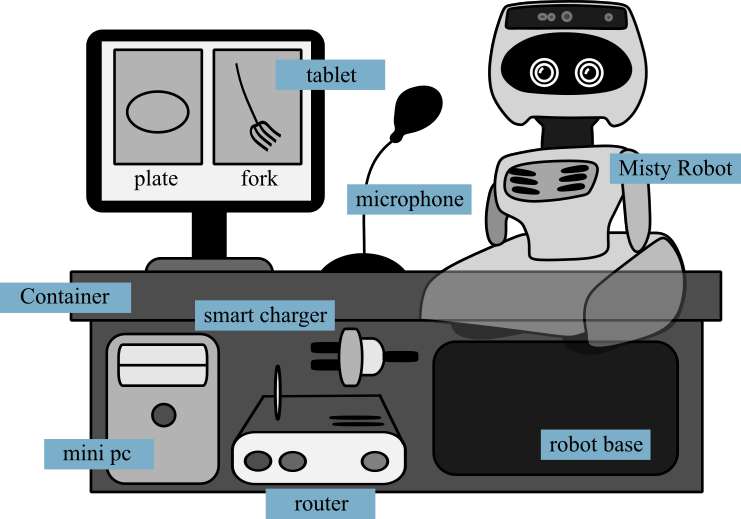}
    \caption{Co-STAR's in-home setup: Misty robot, tablet UI, and tabletop mic; under-table container holds the mini PC, router, and smart charger. Example activity (plate/fork) illustrates multimodal (speech + touch) interaction.}
    \label{fig:system}
\end{figure}

Co-STAR integrates a Misty II robot, a microphone, a mini-CPU, a tablet and a tablet holder (see Figure \ref{fig:system}). The tablet provides visuals for the iCST sessions. A router allows easy connection to a PwD's local internet, allowing Co-STAR to connect to the mini-CPU and to connect our speech-to-text algorithms. The mini-CPU serves as the primary processing unit. We opted against collecting video data to increase the privacy of the PwDs. The components are stored in a sturdy custom enclosure for ease of transportation and user safety. Misty performs gestures while remaining within the enclosure. All data was stored locally within the home.

\subsection{Software Architecture}
The architecture combines speech-based communication with visual assistance and adaptive personalisation. The system includes OpenAI's text-to-speech model for natural voice generation and WhisperD~\cite{akinrintoyo2025whisperd}, a fine-tuned variant of Whisper for speech-to-text. This enabled the system to achieve state-of-the-art performance in handling filler words and disfluencies common in dementia speech. All robot text was pre-generated before deployment. 

A central controller managed session scheduling, activity sequencing, and data flow between system components. It ensured that the session activities were synchronised with the tablet. An Android-based application provided a dual-purpose interface. It provided temporal information by displaying the current day, data and both analogue and digital time when there was no active session. During sessions, it displayed text, images and activity prompts to enhance engagement through multimodal stimulation.

\section{User Study}
We obtained ethical approval for this study (ethics no: {$7091501$}). Recruitment of persons with a clinical diagnosis of dementia was conducted under the expert guidance of dementia care managers of dementia memory clinics and independent living centres. The guidance helped to recruit nine persons with a sufficient level of capacity to provide informed consent. The PwDs' consent was supported by consent from their caregivers (spouse or partner), where applicable. Participants averaged 79.7 years of age; five were female and four were male (Table \ref{tab:participants}).

The robot was deployed in each home for one week, with participants asked to use it daily. On the first day of deployment, the lead researcher was present to facilitate the initial session when possible and to clarify any misconceptions. Participants also received a \say{cheat sheet} outlining system use, session expectations, and Co-STAR’s capabilities and limits.


\begin{table}[t]
  \caption{Participant information: pseudonym, age, past profession, living status, and sessions completed.}
  \label{tab:participants}
  \centering
  \small
  \begin{tabular}{@{}l c c l c c@{}} 
    \toprule
    \textbf{Name} & \textbf{Age} & \textbf{Gender} & \textbf{Profession} & \textbf{Living} & \textbf{Sessions} \\
    \midrule
    Ray        & 83 & M & Geologist          & Son     & 5 \\
    Julia      & 77 & F & Literature officer & Partner & 3 \\
    Georgia    & 78 & F & Doctor             & Alone   & 1 \\
    Esther     & 81 & F & Unknown            & Alone   & 5 \\
    Henry      & 98 & M & Design engineer    & Alone   & 0 \\
    Benjamin   & 59 & M & Sales Assistant            & Partner & 1 \\
    Gideon     & 78 & M & Postal manager     & Partner & 8 \\
    Priscilla  & 94 & F & Nurse              & Alone   & 5 \\
    Martha     & 69 & F & Pharmacist         & Alone   & 3 \\
    \bottomrule
  \end{tabular}
\end{table}

\subsection{Measures}
Our main measurement throughout the study was how many sessions each PwD completed with the robot. To contextualise the findings, we first administered the Demographics Questionnaire, the General Attitudes Towards Robots Scale (GAToRS) questionnaire, and asked questions on the PwDs' familiarity with common technology. At the end of the study, participants completed a System Usability Scale (SUS) to assess their experience with the system. In addition, they participated in a semi-structured interview to gain qualitative insights into usability, acceptance, and perceived value. Where possible, family caregivers completed a questionnaire to provide complementary perspectives on the interactions and their impact on a PwD. 

\section{Results}

A thematic analysis was conducted on the interview transcripts following Braun and Clarke’s six-phase framework. We coded a subset of transcripts, which was subsequently refined to identify emerging themes. We triangulated findings from the interviews, caregiver feedback, interaction logs and recordings to enhance credibility and confirm consistency across data sources. 

\begin{figure*}[t] 
    \centering
    \includegraphics[width=1.0\textwidth]{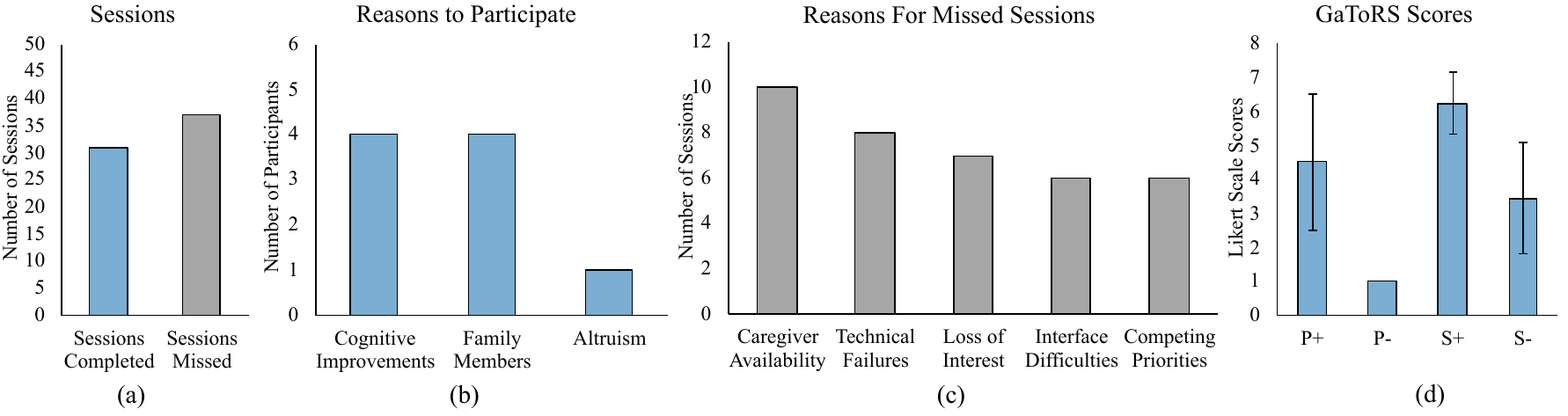} 
    \caption{(a) Number of completed sessions during the $63$-day study period: $31$ sessions completed across $26$ days and $37$ sessions missed, (b) The main reasons for participation were potential cognitive improvements and encouragement from family members, (c) Reasons for missing sessions included caregiver unavailability, technical difficulties and loss of interest, (d) Participants had high positive trust/attitude toward robots (P+) and low negative trust/attitude toward robots (P-). They saw a high positive societal impact of robots (S+), and a moderate negative societal impact of robots (S-).}
    \label{fig:wide}
\end{figure*}

\subsection{Usage}

Participants completed a total of $31$ sessions (Figure \ref{fig:wide}(a)), with an average of $3.44$ sessions per person. Since participants were asked to complete one session per day for one week, this corresponds to roughly half of the expected sessions. Gideon completed the most sessions, with a total of eight, with three of these done on the first day. Three participants (Ray, Esther, and Priscilla) completed five sessions each. The remaining participants completed fewer than half of the sessions, with Henry not completing any. 

On average, participants interacted with the robot for $32m42s$ per session, closely matching the goal of $30$-minute sessions. Georgia completed only one session, which lasted $65m24s$. Priscilla had the highest average session length, with $52m36s$ across her five sessions. Martha recorded the lowest average session length at $23m54s$, excluding Henry, who had an average of $0m$, as he did not complete any sessions. The variation in session length is mainly due to the length of the user responses to the robot.


Participants reported joining the study for three distinct reasons (Figure \ref{fig:wide}(b)). Georgia, Gideon, Priscilla and Esther joined due to the potential benefits of robot-assisted cognitive stimulation therapy. Georgia and Gideon joined after witnessing dementia’s toll on people close to them, including loved ones who had lost their lives to it. Priscilla and Esther hoped a structured, supportive routine could bolster memory. A second group enrolled because of caregiver commitment. Family members of Ray, Julia, Henry and Benjamin sought to help their loved one with dementia. Lastly, Martha joined out of altruism, wanting to be part of dementia research to help others. Four out of nine (Gideon, Benjamin, Georgia, and Esther) participants asked if they could keep the robot for longer at the end of the study. Similarly, the caregivers of Julia and Ray expressed a desire to keep the robotic system for longer than a week. 

These sentiments align with the participants' interviews, which frequently expressed the potential of robot-assisted therapy. As Benjamin put it, \say{I'll use the robot if I learn [new things]}. Similarly, Esther echoed the value of challenge and enjoyment: \say{It made me remember things, which is good for my memory… Some were easy and some that were not easy, but it’s good because it made me remember things}. Julia's partner noted that sessions were \say{\textit{quite good at triggering memories of her childhood, it was relaxing}}. 



Participants reported a variety of reasons for missing sessions (Figure \ref{fig:wide}(c)). The most common cause ($10$ sessions) was caregiver unavailability. Participants relying on caregiver support were especially vulnerable to this, most notably Henry, who was unable to complete any sessions. The second most common cause ($8$ sessions) was technical failures, which arose from issues such as power loss and unstable internet connections. For instance, Benjamin’s internet was unreliable, and Gideon unplugged his system, leading to several missed days. Loss of interest in the system accounted for $7$ missed sessions, reported by Julia and Martha, who eventually stopped using it. Difficulties with the interface were cited as another barrier ($6$ sessions missed). For example, Esther struggled to navigate the tablet and could not use the swiping motion required to unlock it and start a session. Finally, competing life priorities explained the remaining $6$ missed sessions. For example, Georgia was motivated to engage with the robot but prioritised spending time with her visiting grandchildren.



\subsection{Perceptions of the System}

\subsubsection{Quantitative Results}
Six participants provided complete System Usability Scale (SUS) responses. SUS scores were calculated following the standard method, scaled to a range of $0–100$. The average SUS score was \textbf{80.8} ($SD \approx 25.1$; range = 40--100; $n = 6$). This mean score exceeds the established average of $68$, suggesting that the system demonstrates good usability. These statistics exclude Henry (no sessions), Benjamin (travel plans interfered with the interviews), and Martha (conflict in deployment scheduling). 


Eight participants provided GAToRS responses. Priscilla’s responses were not collected due to scheduling constraints. We summarise three informative subscales (1–7 Likert). Positive trust and attitudes towards robots (P+) averaged $4.53$ (SD = $2.01$). Negative trust and attitudes towards robots (P-) averaged $1.00$ (SD = $0.00$). Positive societal impact of robots (S+) was consistently high, with a mean of $6.25$ (SD = $0.90$). Lastly, the negative societal impact of robots (S-) was moderate overall, with a mean of $3.45$ (SD = $1.64$).




\subsubsection{Qualitative Results}
Participants reported enjoying the robotic system, especially when it sparked memories and offered enjoyable challenges. Benjamin set the tone on the first day: \say{\textit{I’ll use the robot if I learn [new things]}}, linking continued use to personal growth. Esther reflected her level of enjoyment: \say{\textit{I like the games a lot, I was laughing}}. Caregivers noticed the same effect. Gideon commented that the activities provide: \say{lots of advantages for the elderly, especially for people with dementia}. Beyond cognition, people appreciated companionship and social ripple effects. This was most notable for Esther. She described that: \say{\textit{I talk to it in the morning and sometimes before I go to bed... I watched a film [at night] and came to it}}. Benjamin's partner signed off the deployment with \say{\textit{Bye-bye, Misty}}, a small but telling attachment cue.

The most consistent criticism was the sense of scriptedness and weak responsiveness. Ray expressed that \say{\textit{It asked questions that are predetermined… because it’s a robot}}. Gideon noted \say{\textit{I expected it to react [to what I said]}}. Such machine-like impressions decreased the perceived value when the robot didn’t acknowledge prior utterances. Also, the content felt repetitive for some (Ray, Julia, Georgia). While Gideon's partner had the same perception, Gideon liked it, noting that: \say{\textit{I don't mind the repeated questions, I do that with my puzzles}}. Others noted the slow system startup and slow responses (Ray, Julia, Priscilla, Martha). Priscilla's daughter noted that while the overall usage was pleasant, she disliked the slow system startup. Martha's disengagement was due to the slow responsiveness of the system.  

The physical look appealed mostly to all participants (\say{\textit{it looks great}} (Gideon)) except Julia. Gideon and Martha expected more capabilities (such as mobility). They wanted a robot capable of moving around their homes. Gideon stated that \say{\textit{If I could move it out of the box, I would have it out walking with me}} while Martha echoed this: \say{\textit{I thought it would move around here [my living room]}}.

\subsection{Social Context and Family Role}

On average, participants who lived with caregivers completed more sessions with a robot (M=$4.3$, SD=$3.0$) than participants who lived alone (M=$2.8$, SD=$2.3$). Qualitative accounts explain this gap: caregivers often initiated and sustained use. They handled accessibility issues with tablet operation. Julia's participation was largely driven by her partner, who reflected: \say{\textit{I was the driving force behind trying it. She wouldn’t have done it on her own}}. Benjamin's involvement was also organised and prompted by his partner. Notably, Priscilla lived alone but relied on her daughter to operate the system. The two days she missed were due to her daughter’s unavailability. Henry also lived alone and completed no sessions because his children visited only occasionally. Ray, Julia, Benjamin, and Priscilla completed all sessions with a caregiver present. Martha completed two out of three sessions with her caregiver. All other sessions were completed independently. However, living with family members also acted as a deterrent for system usage for some participants. Participants such as Ray and Julia often reported lower intrinsic motivation because their existing social networks already provided the stimulation. For instance, Ray expressed limited interest because he regularly interacts with his children, friends and professional caregivers who visit thrice daily. He reflected that his perception \say{\textit{would be a different story if I were living by myself}}.


\subsubsection{Social Validation and Acceptance} 

Several participants expressed curiosity about whether others were using Co-STAR. They sought reassurance through collective adoption. For instance, Georgia asked if others in her group had access to the robot. Julia's partner asked if \say{\textit{others had used it before [his wife]}}. Ray asked if \say{\textit{other people are using it}}. Esther shared her experience with her neighbours, noting \say{\textit{I asked my neighbour to see it}}. Her neighbour responded positively that it “was good” for her. She informed her dementia support networks that the robot was coming. Gideon encouraged others within his community to participate, while they often asked him how he was getting on with Co-STAR. Georgia mentioned that \say{\textit{My ten-year-old… he’s going to be quite excited about it}}.

\subsubsection{Family Bonding and Shared Memories} 
Completing the sessions as a triad also presented opportunities for deep conversations between the PwD and their caregiver, such as revealing previously unknown personal memories. For instance, during the categorisation session, Co-STAR showed a motorbike and asked Ray if he had ever ridden one. He replied, \say{\textit{Yes, I have”}}. Surprised, his daughter responded, \say{\textit{“Really?}} to which Ray said, \say{\textit{Yes, before you were thought of}}. Similarly, Julia shared a childhood memory of riding a motorbike. This was unknown to her husband. Thus, it sparked a rich conversation between them and led to reminiscing about other personal milestones. Likewise, for Julia, the image of a flower reminded her of the flowers her mother gave her that she kept in her garden. This made her reminisce with her partner about memories of her mother. Such moments demonstrate that robot-delivered therapy can foster family bonding. It can prompt storytelling and shared memory reconstruction.

\subsection{Role of Personalisation}
Participants’ responses highlighted that personalisation was vital for sustained engagement.


\subsubsection{Sensory Preferences} Several participants requested adjustments to the system settings. For example, Esther asked that the robot volume be set to a mid-level, as she was sensitive to loud noises. Priscilla's daughter requested a volume control feature. Ray requested the tablet's brightness to be adjusted to a comfortable level to avoid eye strain.

\subsubsection{Adaptive Cognitive Challenge} Participants differed widely in their preferred activity difficulty. This highlights the need for personalised stimulation. Educational background was one of the factors that influenced such preferences. Ray (PhD) found the tasks too simple, stating that \say{\textit{I have quite a lot of intellectual stimulation in my normal life, and something like this could not compete with that}}. He stressed that \say{\textit{For me, it would have to be at a different level. One size fits all is impossible. Each one has to be individually tailored}}. Conversely, Esther was motivated by the more challenging activities. She explained that \say{\textit{The more difficult, the more it helps me...I like to push myself}}.

\subsubsection{Content} Ray's longest responses occurred when Co-STAR asked about places he had lived. He remarked: \say{\textit{This is fascinating, I’ve never done anything quite like this before}}. He stressed that \say{\textit{When it showed me places I’ve been, I was very engaged and interested in chatting about them}}. Personalisation amplified Georgia’s engagement: when Object Categorisation aligned with her cultural knowledge, she reframed the task as teaching: \say{\textit{I could teach it a few things about mangoes because India has about $20$ varieties}}.

\subsubsection{Voice Accent Familiarity} Co-STAR's American accent affected engagement for some participants.
Ray remarked, \say{\textit{It sounds so American}} on six occasions, referring to the speech. Julia also mentioned the American accent during her first session. This suggests that the accent made it less appealing to the British participants. Julia's partner cited the accent as reducing her comfort and trust. He stated that \say{\textit{She found the American accent odd, if you’re going to do a program like this with English people, then it ought to be an English sort of voice}}. In contrast, participants such as Georgia and Esther expressed that they had no concerns regarding the accent.

\section{Case Studies}

We present four case studies of participants who either adopted the technology or did not engage with the system.

\subsection{Gideon – The Consistent User}
Gideon uses a mobile phone, a laptop and an iPad. He is tech-positive: \say{\textit{"I like technology, I really do"}}. He was highly motivated from the outset. He sent multiple emails to confirm the deployment. He lives with his partner but did not need assistance to use the system. 

Gideon exceeded the weekly adherence target with eight sessions ($\approx 274.6$ minutes total; $34.3$ minutes average; $46.1$ minutes longest). He showed strong early momentum with three sessions completed on the first day. He often returned for additional sessions on the same day when convenient. He missed two days because the system was down after he unplugged it. He particularly enjoyed the Faces Game: \say{\textit{there's no correct answer, I think it would suit lots of people because it's [asking for] your opinion}}. However, he was critical of some face choices: \say{\textit{The only thing I was critical about was that I can't stand politicians}}. Also, he wanted more visible responsiveness from Co-STAR: \say{\textit{I thought it would react to it, but it didn't}}. 

Beyond his use, Gideon advocated for the study of the technology and sought to build his community around it. He invited two friends (PwDs) over to his home to watch him do a session. He noted that: \say{\textit{I showed it to my friend [living with dementia] in Australia on a video call}}. He summed up his experience: \say{\textbf{I like the whole thing. It was good}}. He reflected on the broader value: \say{\textit{it's the sort of thing they [people with dementia] need, especially someone like me"}}. He added that: \say{\textit{when the brain fog happens, it controls you, this helps you to get your mind out}}. He concluded that: \say{\textit{lots of advantages for the elderly, especially for people with dementia... people with dementia live in their own world most of the time, this helps you to just focus}}.

\subsection{Georgia - Competing Priorities}
She had no tech difficulties, while also stating that her \say{\textit{ten-year-old grandson could operate it}}. She was highly motivated to use and keep the system, partly due to witnessing the impact of dementia on a close friend she had recently lost. She reported no technical difficulties: \say{\textit{my ten-year-old grandson could operate it}}. However, she only completed one session with the robot. This session lasted about $65$ minutes, due to her detailed responses to the robot. Her reasons for not completing more sessions were due to competing priorities. She discussed her grandchildren who were visiting, a recent move, and belongings still at her previous home. She noted: \say{\textit{I have so many things going on, and thought I could manage this too}}. Additionally, she had assumed the system would stay for longer: \say{\textit{I thought it would stay for longer... thought I could keep it for a month}}. Thus, she asked for it to be available the following month, when her grandchildren returned to school.

\subsection{Benjamin - Limited by Technical Barriers} 
Benjamin spends most of his time watching television. He uses a touchscreen mobile phone for his daily activities, including reminders. He lives with his partner, and they were keen on having him try to use the system. However, she worried that it would be difficult to get him to focus during a session with the robot. Hence, she asked us to reduce the session length to 15 minutes. 

He completed one session lasting approximately $19$ minutes during the study. On the first day, he noted that: \say{\textit{the appetite to learn is not much... I don't use it anymore, I don't speak much}}. However, Benjamin’s interest was not the only problem. Subsequent attempts to complete the sessions failed due to repeated robot disconnections from the home Wi-Fi. While his partner motivated him to participate and engage, she struggled to operate the tablet correctly. Benjamin represents users who are willing but blocked by technological barriers. This highlights the need for seamless integration into their existing network ecosystem.

\subsection{Priscilla — Family-Supported Engagement}
Priscilla is an independent woman with limited technology familiarity. She lives alone and typically leaves her apartment only when accompanied. Her daughter helped her sign up for the study. When the robot arrived, she expressed apprehension: \say{\textit{I’ve not done anything like this before}}. Hence, she discussed participation with her daughter, who assured her that: \say{\textit{I will always do with you}}. Her daughter operated the system and reported that: \say{\textit{I found it absolutely fine}}. Over the study week, Priscilla completed five sessions (total $263.1$ min; average $52.6$ min; longest $66.4$ min). In Faces and Places activities, she did not always recognise people or locations. However, her daughter noted that \say{\textit{doesn't matter if you are doing it with somebody... you [the PwD] could still find something to talk about [with the robot]}}. 

At the end, her daughter described that: \say{\textit{she was worried and nervous [initially]... but then she lost the fear factor}}. However, she noted some usability requests: \say{\textit{it seemed quite slow to load}}, referring to the system boot up. She asked for improvements, such as the addition of a timer to show how far through the session was completed, and a volume control. Her daughter observed improvements in Priscilla’s mood, engagement, companionship, and cognitive stimulation, and summarised: \say{\textit{I found it very helpful as a caregiver}}. Priscilla’s case shows users can go from hesitation to reliable use with minimal caregiver support. Her two missed days were due to her daughter’s absence.


\section{Discussion}

The one-week in-home use of Co-STAR benefited PwDs. 

\subsection{Positive Impact of the System}

Across nine homes, participants completed \textbf{31 sessions} in one week with an average of \textbf{3.44} per person. Using the common iCST benchmark of $\geq 2$ sessions/week, \textbf{6/9} households met the target. Notably, 75\% of PwDs living with family and 60\% of independently living PwDs achieved this benchmark. For a first, unsupervised week, this provides a strong signal that an autonomous, in-home robot can deliver structured cognitive work at a level comparable to human-facilitated baselines. Notably, adherence was often hindered by technical issues rather than motivation, suggesting a higher ceiling with improved reliability.

Co-STAR provided benefits that were cognitive, social and practical in everyday life. On average, the usability was perceived as high (GaToRS S+ M=$4.53$), and four participants asked to keep the system for longer than a week. Participants were informed of future deployments. This indicates acceptability extended past first-use curiosity.  The sessions were described as engaging and enjoyable. The activities matched their interests and abilities. Participants described a blend of cognitive lift and positive affect—an important pairing for sustained use. As Esther put it, \say{\textit{I feel happy\ldots I laugh\ldots I could understand everything\ldots it made me to remember [things]}} (Esther). Patricia's daughter also noted the mood gains she noticed.

Participants and caregivers described sessions as cognitively stimulating and an acceptable challenge. Gideon emphasised reminiscence: \say{\textit{It pushed your brain (...) opening up what you want to be opened up}}. Even when an activity was difficult, the interaction drew out memories and language. Patricia's daughter observed that her mum could: \say{\textit{[mom] could [always] find something to talk about}}. She also noted that it improved Patricia's mood. Several cases illustrated deep reminiscence. Ray lingered over images from the Solomon Islands, recounting detailed work travel experiences, while Julia reflected on childhood after the Second World War. Despite Julia's withdrawal due to her privacy concerns, her partner’s assessment remained positive: \say{\textit{The images and the questions were stimulating, and she was able to say quite a bit, it could be a fantastic tool}}. Combined, these accounts show how Co-STAR facilitated meaningful cognitive engagement, regular practice, and mood improvement.



\subsection{The Role of Family Members}

Notably, we found that therapy at home is mostly triadic, not dyadic. \say{Living alone} did not guarantee independent use. Caregiver unavailability coincided with missed therapy days (10 sessions). Henry, whose children visited infrequently, completed no sessions. Dyads generally had better adherence. Yet, they were fragile to caregiver schedules and infrastructure faults (e.g., Benjamin’s repeated Wi-Fi disconnections despite partner prompting). In practice, sessions were a negotiation among the person, the caregiver and the robot. Caregivers brokered trust, supported setup, provided nudges and often co-answered. Moreover, co-presence often shifted a hesitant user into participation. 

The path forward for HRI is clear. Based on the findings of this work, home iCST robots should: (1) provide two explicit modes (autonomous and caregiver-assisted) with on-screen role/turn prompts; and (2) support multi-user turn-taking and joint decisions by default. Designing for independence and partnership (acknowledging the triadic reality) makes iCST feasible and accessible for those living alone while supporting dyads.

\subsection{Trust, Privacy, and Perceived Competence}
Across the deployments, a frequently mentioned issue was whether interactions felt comfortable and trustworthy. Our usability data support this reading: although the mean SUS was \textbf{80.8}, scores varied substantially (range \textbf{40--100}). Some had moderate worries about wider impacts (GaToRS S- was M=$3.45$). This variation mapped to how safe, responsive, and respectful the system was perceived to be, rather than to generic beliefs about robots. Participants put it plainly: Henry was \say{\textit{not sure}} he could trust a robot because he had \say{\textit{not experienced one}}; Ray began with \say{\textit{No. Definitely not}} (and \say{\textit{Not entirely}} regarding developers) and added, \say{\textit{I'm very skeptical of talking to robots}}, yet was willing to recalibrate his trust—\say{\textit{I might be wrong\ldots I'm still a scientist. I'm always prepared to change my mind in the face of new evidence}}. Julia's hesitation centred on privacy—\say{\textit{Is my name going to be everywhere?}}. When Co-STAR's responses felt scripted or non-contingent, perceived competence dropped—\say{\textit{It can't think\ldots it's a machine}} (Ray). Practically, future therapy providing robots should prioritise responsiveness over scripted exchanges, reduce latency, provide continuous personalisation, and make privacy-by-design explicit. Enhancing trust through these can increase use.




\subsection{Limitations and Future Work}
This study highlights the potential of SARs to support cognitive therapy for PwDs. Future work will involve a long-term deployment for several months with a larger sample size to track cognitive changes and assess long-term benefits.

\bibliographystyle{IEEEtran}
\bibliography{IEEEabrv,bibliography}


\end{document}